\documentclass[11pt,a4paper]{article}
\usepackage[hyperref]{acl2019}
\usepackage{times}
\usepackage{latexsym}

\usepackage{url}
\usepackage{CJKutf8}
\usepackage{graphicx}
\usepackage{subfigure}
\usepackage{multirow}
\usepackage{amssymb}
\usepackage{array}
\usepackage{booktabs}
\usepackage{enumitem}

\usepackage{url}

\setitemize[1]{itemsep=3pt,partopsep=0pt,parsep=\parskip,topsep=5pt}
\aclfinalcopy 

\title{Zero-shot Chinese Discourse Dependency Parsing via Cross-lingual Mapping}

\author{
  Yi Cheng \qquad Sujian Li\\
  MOE Key Lab of Computational Linguistics, School of EECS, Peking University\\
  Peng Cheng Laboratory, Shenzhen, China\\
  {\tt \{yicheng,lisujian\}@pku.edu.cn}\\}

\date{}

\begin{document}
\maketitle
\begin{abstract}
Due to the absence of labeled data, discourse parsing still remains challenging in some languages. In this paper, we present a simple and efficient method to conduct zero-shot Chinese text-level dependency parsing by leveraging English discourse labeled data and parsing techniques. We first construct the Chinese-English mapping from the level of sentence and elementary discourse unit (EDU), and then exploit the parsing results of the corresponding English translations to obtain the discourse trees for the Chinese text. This method can automatically conduct Chinese discourse parsing, with no need of a large scale of Chinese labeled data.
\end{abstract}

\section{Introduction}
Discourse parsing aims to analyze the inner structure of texts, which is fundamental to many natural language processing applications, such as question answering and summarization. 
The construction of discourse corpora has promoted the development of discourse parsing techniques.
In English, the widely-used discourse corpora include the Rhetorical Structure Theory Treebank (RST-DT)~\cite{Carlson2001} and Penn Discourse TreeBank (PDTB)~\cite{PDTB2008}.


Recently, \newcite{Li14} and \newcite{Yoshida2014} proposed the discourse dependency structure (DDS). DDS directly links the EDUs, so it has fewer nodes and simpler structures compared to RST and PDTB. In addition, it can easily represent non-projective structures, while hierarchical structures need other complex mechanisms to do so. DDS is especially important for Chinese. \newcite{Kang19} analyzes almost all the existing Chinese discourse treebanks and concludes that DDS is the future direction due to its right balance between expressiveness and practicality. 
However, little research has been done on Chinese DDS. On one hand, there have been no such DDS treebanks in Chinese yet. Most of the existing Chinese discourse corpora follow PDTB-style or RST-style annotation \cite{Zhou2012,Zhou2015,Yue2008}. Building a high-quality DDS corpus from scratch is labor-intensive and there are some conversion problems in transforming an existing corpus into DDS. On the other hand, a Chinese discourse parser needs to explore efficient features through trial and error based on the characteristics of Chinese. For the above reasons, Chinese text-level dependency parsing remains challenging.

To overcome these problems, 
we propose a simple and efficient method that conducts zero-shot Chinese discourse dependency parsing by exploiting the existing English discourse resources, 
with no need for Chinese training data. 
This is motivated by the observation of some Chinese-English parallel sentences such as the examples in Fig.1, whose dependency parsing trees are the same.
It can be seen from the figure that the logical organization of a text is similar at the macro discourse level regardless of languages, 
in spite of lexical or grammatical  differences.

Based on this observation, we employ machine translation (MT)  and English discourse parsing techniques to parse a Chinese text. 
Our proposed method is simple but feasible,
because English discourse dependency parsing has made progress, especially in parsing discourse tree structures \cite{Yang2017,Kim2017}, and 
Chinese-to-English MT techniques are relatively mature
\cite{Translation1,Translation2}.
Specifically, we first make use of MT techniques to translate a Chinese text into English and then adopt a transition-based English parser to analyze the translated text. 
Finally, we map this English parsing result to the Chinese text. During this process, some modifications are made to MT and the parsing result for performance improvement .

\begin{figure}[t]
  \centering
  \includegraphics[width=\linewidth]{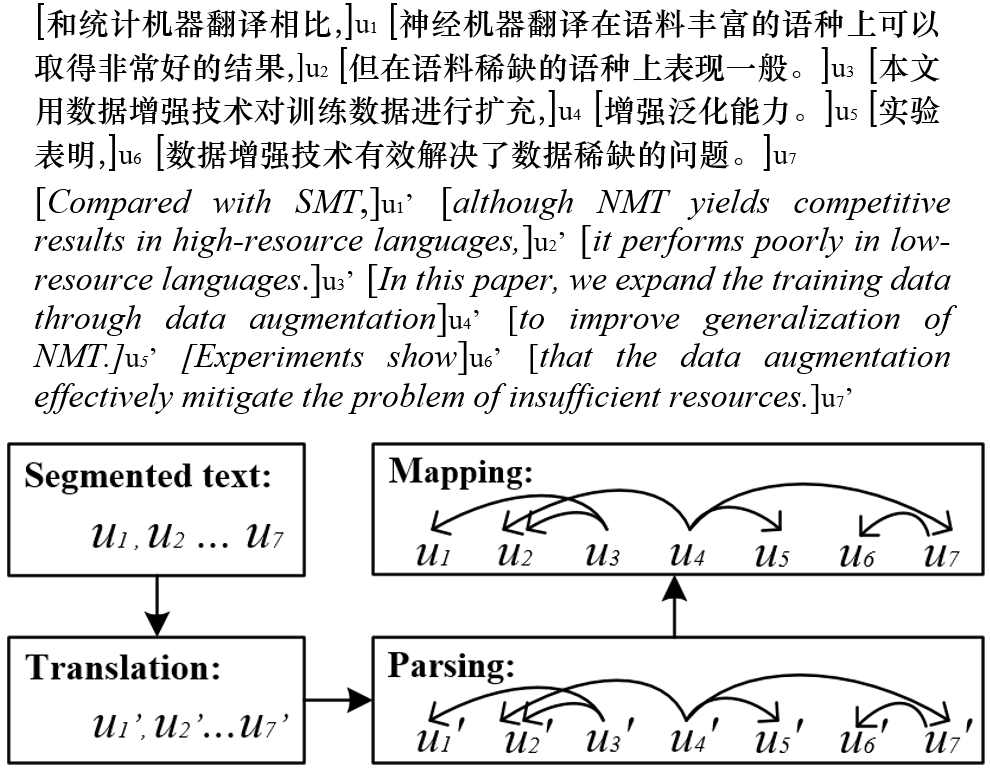}
  \caption{Illustration of Our Parsing Method via a Chinese-English Parallel Example}
  \label{flowchart}
\end{figure}

To evaluate our proposed method, we manually construct a small dataset, on which our method exhibits promising performance. This corpus will be released soon. The experiment results demonstrates that our method is potentially helpful in building large-scale  data for Chinese neural NLG systems that make use of discourse structure.
To the best of our knowledge, we are the first to conduct discourse dependency parsing in Chinese. 

\section{Chinese Discourse Dependency Corpus Construction}

In this work, a small-scale Chinese discourse dependency treebank is constructed for evaluation. Here, we primarily follow the guideline of building the English discourse dependency treebank SciDTB~\cite{Yang2018} to explore the specifics of labeling DDS in Chinese.

First, 
scientific abstracts are chosen as corpus sources,
because they are short texts with obvious logic and within the same domain as 
the English treebank (SciDTB) \cite{Yang2018}.
Specifically, 108 abstracts are selected from a Chinese NLP journal \emph{JCIP}
\footnote{http://jcip.cipsc.org.cn/CN/volumn/home.shtml}.

Second, we manually separate these abstracts into elementary discourse units (EDUs),
the basic units of a parsing tree. Each segmented abstract is checked at least twice to ensure segmentation quality. 
Our EDU segmentation mainly refer to the criteria of RST-DT~\cite{manual} and make some modifications to the guideline based on the linguistic characteristics of Chinese \cite{Cao2017,YangJf2018}.
Due to space limitation, we do not list these modifications, as 
EDU segmentation is not the main work of this paper. 

Third, for each abstract, we identify the head of every EDU and the relation type between them, which is the most labor-intensive of all steps.
We adopt the head and relation identification guidelines defined in \newcite{Yang2018}. The relation categories include 17 coarse-grained and 26 fine-grained relation types.
During the annotation process, some relation types are hard to distinguish (e.g., the distinction between the relations ``manner-means'' and ``enablement'' is vague). 
In addition, relation pronouns (e.g., that) and conjunctions (e.g., but) are used less frequently in Chinese
\cite{Li2014}, adding to the difficulty of relation labeling. 
The primary target of this study is to automate this step, i.e., to build the discourse tree with relation types between EDUs identified for a Chinese text.

\begin{table}[!tb]
   \centering
   \begin{tabular}{ccc}
     \toprule[1pt]
     \textbf{Relation} & \textbf{Frequency} & \textbf{Percentage}/\% \\
     \bottomrule[1pt]
     elab-addition & 408 & 29.31\\
     joint & 236 & 16.95\\
     enablement & 138 & 9.91\\
     bg-general & 135 & 9.70\\
     evaluation & 85 & 6.10\\
     \bottomrule[1pt]
   \end{tabular}
   \caption{The Most Frequent Relation Types}
   \label{Relation}
\end{table}

Two annotators first learned the annotating principle before the annotation work. It takes the annotators 3 months to label the 108 abstracts, each being labeled at least twice independently in order to check annotation consistency and provide human performance as an upper bound. 30 abstracts are used for validation and the rest for test.
The inter-annotator agreement is 0.780 and 0.673 with respect to UAS and LAS.
In total, there are 1,500 EDUs 
(including 108 artificial root EDUs) 
with an average of 12.9 EDUs per abstract
and 1,392 labeled discourse relations.
On average, there are 2.91 EDUs per sentence and 22.17 characters per EDU. 
Table \ref{Relation} shows the five most frequent relation types, along with their frequencies.

\section{Zero-shot Chinese Dependency Parsing}

As stated above, our method aims to generate a
dependency parsing tree with relation types between EDUs identified for a Chinese text.
It is assumed that golden EDU segmentation has already been conducted for the text. 
Formally, 
given a Chinese text $t_{C}$=
$(u_{1},u_{2},...,u_{k})$ composed of $k$ EDUs,  
we translate each Chinese EDU $u_{i}$ directly into  $u_{i}'(i$=$1,2,..k)$, which can be seen as an English EDU.
Translation performance is restrained to a certain extent because some EDUs cannot individually express their precise meaning when taken out of context. Thus, we make some modifications to the translation results before adopting a transition-based parser to generate a discourse dependency tree for the translated English EDUs. Finally, this dependency tree is mapped onto the EDU-segmented Chinese text.
Fig.\ref{flowchart} illustrates the whole process of our method.
The main idea is simple. Only some technical issues in translation and text parsing need addressing, 
which will be introduced in the subsections.

\subsection{Translation}
We translate each Chinese EDU separately, 
instead of processing the whole text at a time,
in order to obtain one-to-one correspondence between translated English EDUs and their Chinese counterparts
, and to bypass EDU segmentation in English. 
But due to the absence of context information, 
translation accuracy is sacrificed, which degrades parsing performance. 
Since our work does not involve improving translation techniques, we only modify some obvious translation problems. 

First, in translation, Chinese EDUs with incomplete meaning may be mistranslated into a sentence ended with a period. As \newcite{Zhou2015} point out, punctuation marks in Chinese can serve as clues of discourse relations. Most competitive Chiense discourse parsing models\cite{Kang16} use punctuation as one of their features. Therefore, we stipulate that the translated English EDUs can only be ended with a period if its corresponding Chinese EDU is ended with one. The other periods in the translation are replaced with commas.

Second, we modify the EDU identification of some relative pronouns because the position of them is helpful information for judging specific relation types (e.g.,``attribution''). 
Since we use EDUs as translation units
, the EDU identification of some relation pronouns violates English EDU segmentation criteria. 
Take $u_{6}$ and $u_{7}$ in Fig.\ref{flowchart} as example. Their translations are respectively: \textit{[Experiments show that]$\bar{u}'_{6}$}, \textit{[the data augmentation effectively mitigate the problem of insufficient resources.]}{$\bar{u}'_{7}$}.
Our modification is to move  ``\textit{that}'' from $\bar{u}'_{6}$ to $\bar{u}'_{7}$, because a relative pronoun should be with the clause it introduces, according to the EDU segmentation criteria of RST-DT.

\subsection{English Discourse Parsing}
We follow the work of \newcite{Yang2018} and implement a two-stage transition-based dependency parser based on the idea of \newcite{Wang2017} to conduct English parsing.
In the first stage, the transition-based method for dependency parsing \cite{Nivre2003} is adopted to identify the head for each EDU. 
We employ the action set of arc-standard system \cite{Nivre2004}, and an SVM classifier is designed to predict the most possible transition action. 
In the second stage, another SVM classifier is trained to predict relation types. 

Since this parser is trained with SciDTB, its performance heavily relies on the features of the corpus. 
By analyzing the parsing results on the validation data, we find one obvious problem: the parser identifies the topic EDU (whose head is the root EDU, such as $u_{4}$ in Fig.1) with an accuracy of only 44.95\%, while it reaches  85.06\% on SciDTB.

To alleviate this problem, we first identify the topic sentence (which includes the topic EDU) in a rule-based way, because it usually begins with certain words, such as {\small{\begin{CJK*}{UTF8}{gbsn} ``该文''\end{CJK*}}}(this paper).
Next, we split the passage into two parts with the topic sentence being the beginning of the latter part. 
The two parts are then parsed separately and joined together.  
In this way, the topic EDU identification accuracy increases to 68.52\%.

\section{Experiment}
\subsection{Setup}
In our work, we compared several ready-made translation tools and chose to use \emph{Youdao Translator}\footnote{http://fanyi.youdao.com/}.
We referred to \newcite{Yang2018}'s work  and implemented a two-stage transition-based discourse dependency parser to parse the English translated EDUs, with SciDTB as the training corpus.
For comparison, we adopted the metrics of unlabeled and labeled attachment scores (UAS and LAS).
UAS measures the accuracy of labeling the heads, while 
LAS measures the accuracy with respect to both head and relation labeling.

\subsection{Results}
Since there is no previous research on Chinese text-level dependency parsing, and our parsing approach is mainly designed to help construct a large-scale discourse dependency corpus in Chinese, our major concern is what performance this method (named \textit{Zero-shot} in Table \ref{Performance Comparison}) can achieve and how it compares to human performance. 
We list several parsing results for comparison:

\begin{itemize}[leftmargin=*]
\item \textit{Random} is a transition-based dependency parser which randomly chooses ``shift'' or ``reduce'' as its next action and always uses the most frequent relation type ``elab-addition'' as the relation label.  We test it on our Chinese corpus.
\item \textit{Supervised(Chinese)} is a two-stage transition-based dependency parser trained with 80 abstracts of our Chinese corpus and tested with the remaining 28 abstracts.
\item \textit{Supervised(English)} is a two-stage transition-based dependency parser trained on the training set of SciDTB and evaluated on its test set. 
\item \textit{Human(Chinese)} and \textit{Human(English)} are human performance on our Chinese discourse corpus and SciDTB respectively.
\end{itemize}

 \begin{table}[!t]
   \centering
   \begin{tabular}{ccc}
     \toprule[1pt]
     \ &\textbf{UAS}&\textbf{LAS}\\
     \bottomrule[1pt]
     Random  & 0.188 & 0.013\\
     Supervised(Chinese) & 0.525 & 0.276 \\
    \textbf{Zero-shot}&\textbf{0.643}&\textbf{0.384}\\
     Human(Chinese)& 0.780 & 0.673\\
     \hline
     Supervised (English)&0.702&0.545\\
     Human(English)&0.806&0.627\\
     \bottomrule[1pt]
   \end{tabular}
   \caption{ Performance Comparison with Other Parsing Models }
   \label{Performance Comparison}
 \end{table} 

Table \ref{Performance Comparison}  shows the UAS and LAS of different parsing results. The top four rows are performance tested on our Chinese corpus and the bottom two on SciDTB. From \textit{Human(English)} and \textit{Human(Chinese)}, we can see that discourse labeling is a difficult task for both languages.
Our \textit{Zero-shot} method significantly outperforms the \textit{Random} parser, meaning that parallel English and Chinese texts have similar discourse structures, and that our method effectively leverages such information.
\textit{Zero-shot} also performs about 12\% and 11\% higher than the \textit{Supervised(Chinese)} with respect to the UAS and LAS metrics, because our corpus is too small to support supervision well enough. Compared with \textit{Supervised(English)}, the performance of \textit{Zero-shot} is acceptable in terms of identifying the head EDU, but barely satisfactory in labeling the relations, which might be explained by different statistical distributions of relations types in Chinese and English..

\begin{figure}[t]
  \centering
  \includegraphics[width=\linewidth]{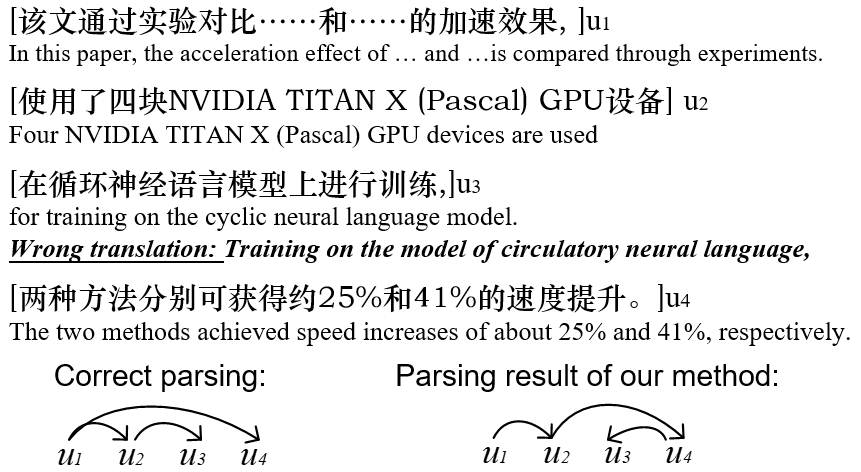}
  \caption{Parsing Errors Caused by Wrong Translation}
  \label{BadExp}
\end{figure}

\begin{table}[!t]
   \centering
   \begin{tabular}{ccc}
     \toprule[1pt]
     \textbf{Ablation test} &\textbf{UAS}&\textbf{LAS}\\
     \bottomrule[1pt]
     Direct parsing
     & 0.500 & 0.312 \\
     + Relative Pronoun Adjustment & 0.527 & 0.333 \\
     + Punctuation Modification & 0.607 & 0.353\\
     + Two-part Parsing & 0.643 & 0.384\\
     \bottomrule[1pt]
   \end{tabular}
   \caption{Ablation Study}
   \label{Abl}
\end{table}

To evaluate the contribution of each modification mentioned in Section 3, we conduct ablation experiments as shown in Table \ref{Abl}. 
The first line displays the performance of direct parsing without any modifications. The next three lines shows the performance with the modification strategies added in turn. 
As demonstrated in the table, these subtle modifications all play a useful role in improving performance.

Through error analysis, we find that many wrong cases can be corrected if the parser is given precise translation. Fig.\ref{BadExp} provides an example where the heads of some EDUs are wrongly labeled, but are correct if given right translation. Translation precision can be improved with consideration of a larger context than EDU, which will be our future work.

 \section{Conclusions}
 In this paper,
 we present a simple and efficient method to conduct zero-shot Chinese discourse parsing, whose performance is close to the one of the state-of-art English parsers.
 It opens the possibilities for conducting dependency parsing on low-resource languages via cross-lingual mapping, reducing human labor of corpus construction.
 In the future, we will further improve our method and test it in more languages and more domains.

\section*{Acknowledgments}
We thank the anonymous reviewers for their helpful comments on this paper. This work was partially supported by National Natural Science Foundation of China (61572049 and 61876009). 
\\

\bibliography{acl2019}
\bibliographystyle{acl_natbib}

\end{document}